\documentclass[10pt,twocolumn,letterpaper]{article}

\usepackage[pagenumbers]{iccv} 

\usepackage[ruled,linesnumbered]{algorithm2e}
\usepackage{algorithmic}
\usepackage{setspace}
\usepackage{multirow}
\usepackage{diagbox}
\usepackage{colortbl}
\usepackage{pifont}
\usepackage[dvipsnames]{xcolor}
\definecolor{iccvblue}{rgb}{0.21,0.49,0.74}
\definecolor{elegantblue}{RGB}{213, 228, 242}
\usepackage[pagebackref,breaklinks,colorlinks,allcolors=iccvblue]{hyperref}

\def\paperID{7883} 
\def\confName{ICCV}
\def\confYear{2025}

\title{AdvDreamer Unveils: \\ Are Vision-Language Models Truly Ready for Real-World 3D Variations?}

\author{
Shouwei Ruan$^{1\dagger}$, Hanqing Liu$^{1\dagger}$, Yao Huang$^{1\dagger}$, Xiaoqi Wang$^{1}$, Caixin Kang$^{1}$, \\Hang Su$^{2}$, Yinpeng Dong$^{2}$, Xingxing Wei$^{1}$\thanks{Corresponding author.} \\
$^{1}$Institute of Artificial Intelligence, Beihang University \\ $^{2}$ Dept. of Comp. Sci. and Tech., Institute for AI, Tsinghua-Bosch Joint ML Center, \\THBI Lab, BNRist Center, Tsinghua University  \\ 
\scriptsize{\texttt{\{shouweiruan, hqliu, y\_huang, zy2342127, caixinkang, xxwei\}@buaa.edu.cn, \{suhangss, dongyinpeng\}@tsinghua.edu.cn}}
}

\begin{document}

\twocolumn[{%
\renewcommand\twocolumn[1][]{#1}%
\maketitle
\vspace{-1.0cm}
\begin{center}
\centering
\includegraphics[height=6.8cm]{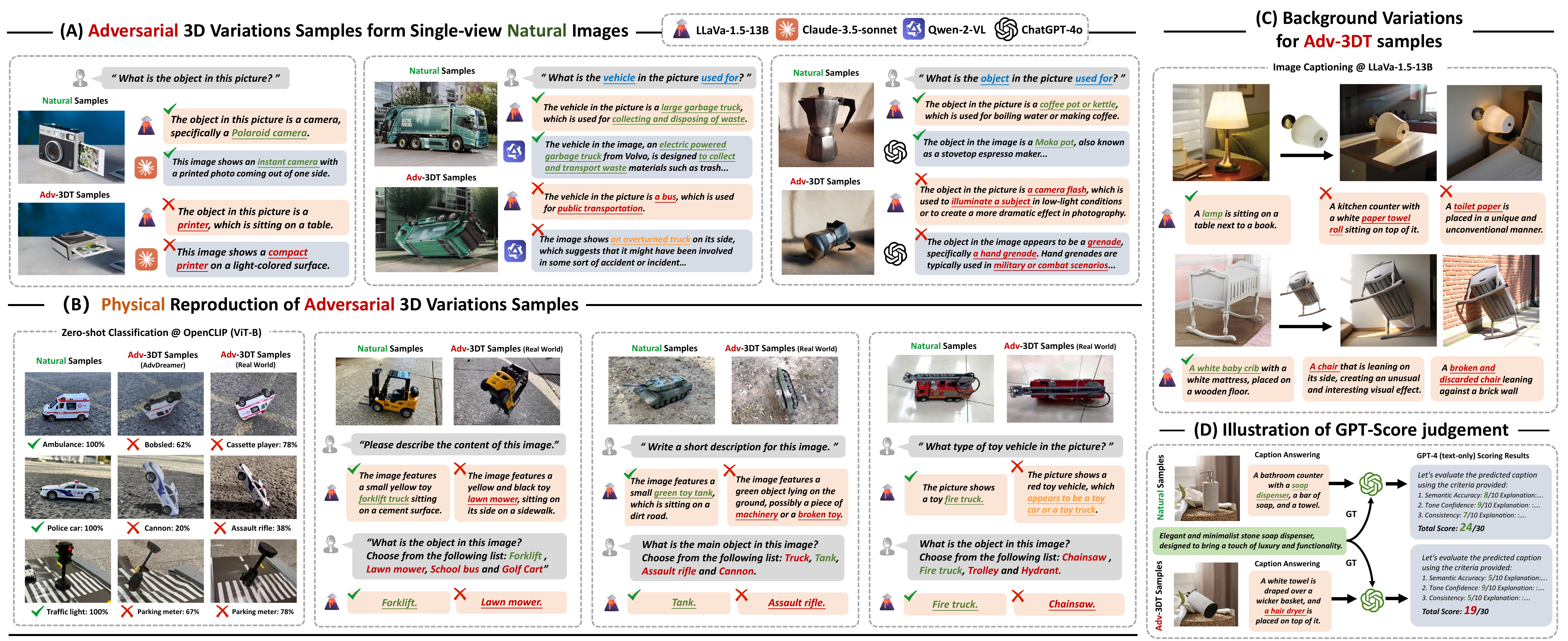}
\captionof{figure}{\textbf{The 3D Variation Vulnerabilities in VLMs}. \textbf{\emph{(A)}} The proposed AdvDreamer allow the capture of adversarial 3D transformation (Adv-3DT) samples from single natural images, threatening various VLMs. \textbf{\emph{(B)}} These Adv-3DT samples can be successfully reproduced in the physical world and \textbf{\emph{(C)}} retain their aggressiveness under background variations. \textbf{\emph{(D)}} We also explain the GPT-Score metric used in the experiments. In each VLM response, we highlight the  \textbf{\textcolor{Green}{correct}}, \textbf{\textcolor{Maroon}{significant error}}, and \textbf{\textcolor{Dandelion}{less precise}} parts using different colors.}
\label{fig:vis}
\end{center}%
}]


\renewcommand{\thefootnote}{}
\footnotetext{$\dagger$ Equal contribution.}
\renewcommand{\thefootnote}{\arabic{footnote}}
\begin{abstract}
Vision Language Models (VLMs) have exhibited remarkable generalization capabilities, yet their robustness in dynamic real-world scenarios remains largely unexplored. To systematically evaluate VLMs' robustness to real-world 3D variations, we propose \textbf{AdvDreamer}, the first framework capable of generating physically reproducible Adversarial 3D Transformation (Adv-3DT) samples from single-view observations. In AdvDreamer, we integrate three key innovations: Firstly, to characterize real-world 3D variations with limited prior knowledge precisely, we design a zero-shot \textbf{Monocular Pose Manipulation} pipeline built upon generative 3D priors. Secondly, to ensure the visual quality of worst-case Adv-3DT samples, we propose a \textbf{Naturalness Reward Model} that provides continuous naturalness regularization during adversarial optimization, effectively preventing convergence to hallucinated or unnatural elements. Thirdly, to enable systematic evaluation across diverse VLM architectures and visual-language tasks, we introduce the \textbf{Inverse Semantic Probability} loss as the adversarial optimization objective, which solely operates in the fundamental visual-textual alignment space. Based on the captured Adv-3DT samples with high aggressiveness and transferability, we establish MM3DTBench, the first VQA benchmark dataset tailored to evaluate VLM robustness under challenging 3D variations. Extensive evaluations of representative VLMs with varying architectures reveal that real-world 3D variations can pose severe threats to model performance across various tasks.

\end{abstract}
\vspace{-0.3cm}    
\begin{figure*}[t]
  \centering
  \includegraphics[height=3.0cm]{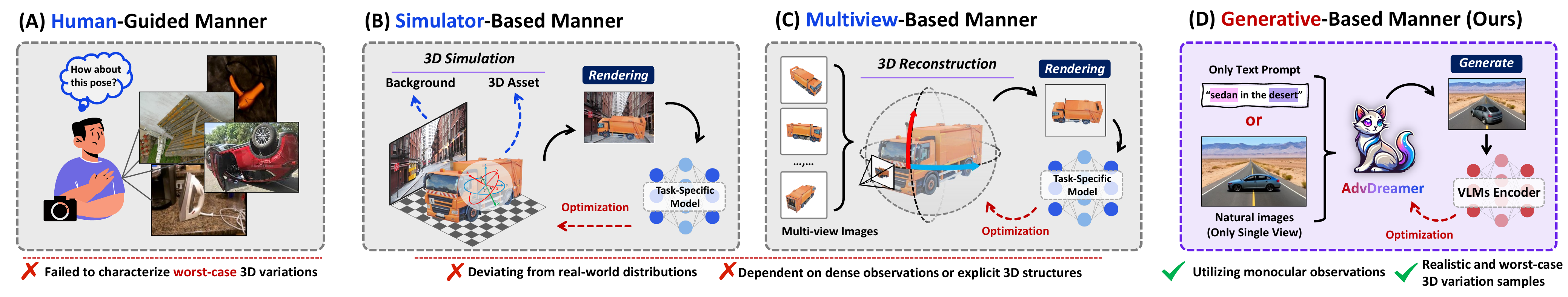}
  \vspace{-0.2cm}
  \caption{Comparison of paradigms to characterize Adversarial 3D Transformation (Adv-3DT) samples.}
  \vspace{-0.5cm}
  \label{fig:intro}
\end{figure*}
\vspace{-0.2cm}
\section{Introduction}
\label{sec:intro}
Recently, Vision-Language Models (VLMs)~\cite{liu2024visual,liu2024llavanext,bai2023qwen,chen2024internvl} have demonstrated remarkable capabilities in bridging visual perception and natural language understanding. Through large-scale image-text pre-training followed by fine-grained instruction tuning~\cite{liu2024visual,li2023blip}, these models effectively tackle a wide range of visual-centric tasks, including instruction-based image recognition~\cite{radford2021learning,zhai2023sigmoid,sun2023eva}, visual question answering~\cite{alayrac2022flamingo,liu2024visual,liu2024llavanext,li2023blip}, and visual reasoning~\cite{chen2024large,chen2024spatialvlm}. Due to their strong generalizability and impressive zero-shot capability, VLMs are increasingly deployed in various safety-critical applications, particularly in autonomous driving~\cite{tian2024drivevlm,xu2024drivegpt4} and robotic systems~\cite{huang2023voxposer,yokoyama2024vlfm}.


Despite extensive studies validating VLMs' resilience against various distribution shifts (\eg, style variations~\cite{zhang2024lapt,al2024unibench}, image corruptions~\cite{al2024unibench}, and adversarial perturbations~\cite{cui2024robustness, zhang2024benchmarking, zhao2023evaluate}), these investigations primarily focus on 2D perturbations in the digital domain, overlooking a critical challenge in real-world deployment: the 3D variations. As VLMs are increasingly integrated into dynamic environments, their ability to handle 3D variation becomes crucial, raising a fundamental question: \emph{\textbf{Have current VLMs attained sufficient robustness to the distribution shifts arising from ubiquitous 3D variations in the real world?}}

To thoroughly explore this problem and offer comprehensive answers, we identify three fundamental challenges: 

\emph{\textbf{(1) How to accurately characterize real-world 3D variations with limited prior knowledge?}} Existing methods exploring the 3D variation robustness heavily rely on explicit 3D structures~\cite{alcorn2019strike, hamdi2020towards} or representations derived from dense multi-view observations~\cite{alcorn2019strike, hamdi2020towards, madan2022and}. However, in most practical scenarios, the available priors are severely limited (\eg, single-view observation only). This constraint necessitates developing a method that minimizes dependency on extensive scene priors. To meet this, we introduce a zero-shot \textbf{Monocular Pose-Manipulation (MPM)} pipeline based on generative representations. Specifically, MPM leverages the rich 3D priors embedded in pre-trained Large-Reconstructive Models (LRM)~\cite{hong2023lrm,openlrm} and Image Re-Composition Models~\cite{chen2024anydoor,zhang2025scaling}, enabling fine-grained 3D adjustments with only single-view images and specified 3D variation parameters. While this generative-based paradigm establishes a foundation for capturing worst-case 3D variations through optimizing 3D variation parameters, it still exhibits inherent limitations that may introduce undesirable image corruptions, thus presenting our second challenge:


\emph{\textbf{(2) How to capture the worst-case adversarial 3D transformation (Adv-3DT) samples while ensuring that the performance degradation is inherently caused by the 3D variation?}} Prior works utilizing synthetic 3D assets~\cite{alcorn2019strike, hamdi2020towards, madan2022and} frequently produce samples with noticeable texture discrepancies compared to real-world images. These limitations in samples' visual fidelity and contextual coherence highlight a critical concern: observed performance degradation under Adv-3DT samples may stem from undesirable corruption rather than the 3D variation itself. To address this, we incorporate naturalness reward constraints within the adversarial optimization process. Specifically, we treat rigid 3D variations as an adversarial attack and propose the \textbf{Naturalness Reward Model (NRM)} that preserves image quality throughout optimization. NRM leverages visual context features provided by the DINOv2~\cite{oquab2023dinov2} backbone to predict a naturalness reward that jointly reflects the visual fidelity and physical plausibility of the samples, effectively regularizing the optimization trajectory and preventing convergence toward unnatural or hallucinated elements.


\emph{\textbf{(3) How to ensure effective generalization of Adv-3DT samples across diverse VLM tasks and architectures?}} Existing studies~\cite{dong2022viewfool, alcorn2019strike, ruan2023towards} typically rely on task-specific objectives with end-to-end optimization, substantially limiting their applicability across diverse VLM architectures and a wide range of visual language tasks. To transcend this limitation, we formulate the adversarial objective in a task- and architecture-agnostic manner. Specifically, we introduce the \textbf{Inverse Semantic Probability (ISP)} loss, which operates solely through the foundational component of the VLM--the visual encoder, aiming to minimize the image-text matching probability between Adv-3DT samples and their corresponding ground-truth textual descriptions. Extensive experiments demonstrate that Adv-3DT samples optimized under the ISP loss exhibit remarkable adversarial transferability across various downstream tasks and models. Building on this, we further introduce the \textbf{MM3DTBench}, the first VQA benchmark dataset specifically designed to evaluate the 3D variation robustness of VLMs in real-world scenarios, comprising highly challenging Adv-3DT samples with carefully crafted candidate answers.

Integrating the aforementioned innovations, this paper proposes \underline{\textbf{AdvDreamer}},  a novel framework that enables the characterization of Adv-3DT samples from monocular observations, with broad adaptability across various visual-language tasks and VLM architectures. Our extensive digital and physical experiments reveal significant vulnerabilities in contemporary VLMs when confronted with challenging 3D variations. In visual recognition tasks, foundational models like OpenCLIP~\cite{openclip} and BLIP-2~\cite{li2023blip} suffer a performance degradation of \textbf{\emph{65\%$\sim$80\%}} on Adv-3DT samples. More concerning still, in open-ended tasks including image captioning and visual question answering, even the most advanced commercial models like GPT-4o~\cite{gpt-4o} exhibit performance declines approaching \textbf{\emph{50\%}}. Our findings highlight the urgent need to enhance VLM robustness against 3D variations and set new requirements for deploying VLMs in complex, dynamic real-world applications.

\begin{figure*}[t]
  \centering
  \includegraphics[height=5.7cm]{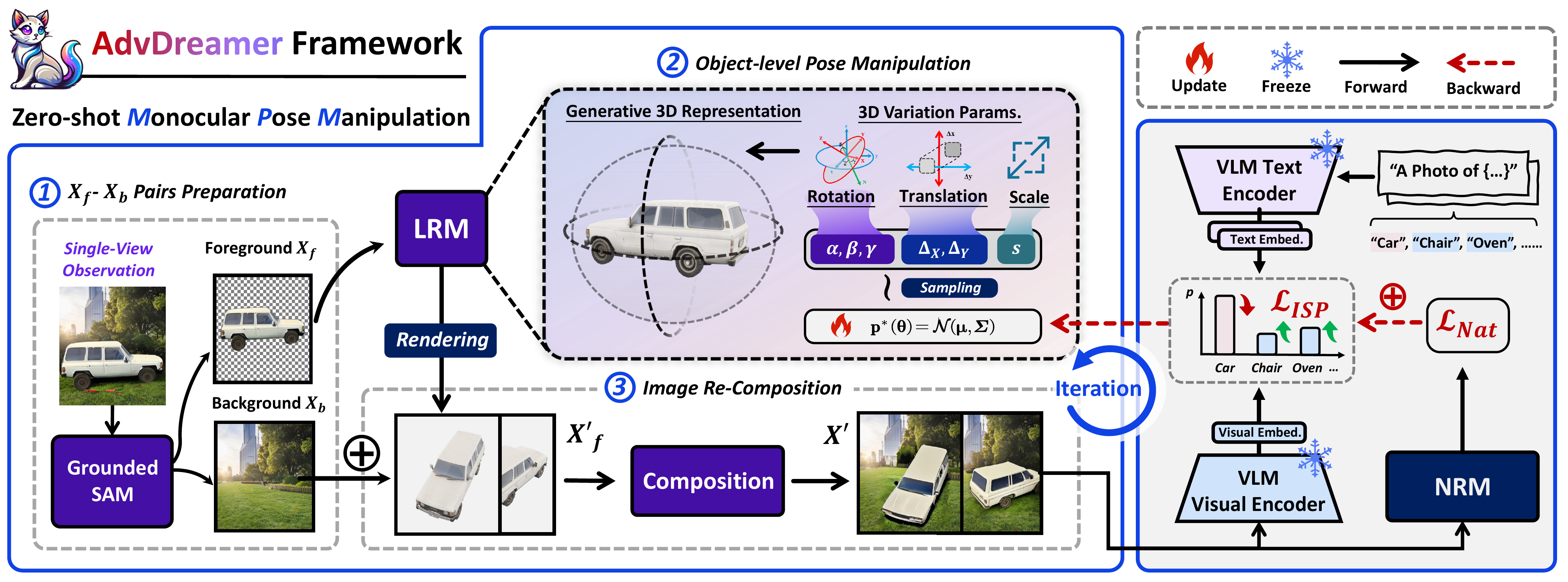}
    \vspace{-0.2cm}
    \caption{\textbf{Overview of the proposed AdvDreamer framework.} To capture worst-case 3D variations in the real world, we treat rigid 3D transformations as adversarial attacks, optimizing the distribution of transformation parameters through a query-based approach. In AdvDreamer, we introduce the Monocular Pose Manipulation pipeline to perform specified 3D transformations on single-view images and guide the optimization process using the proposed Naturalness Reward Model and Inverse Semantic Probability loss.}
  \vspace{-0.4cm}
  \label{fig:framework}
\end{figure*}
\section{Related Work}
\label{sec:Related Work}

\subsection{Vision-Language Models}
Classic Vision-Language Models (VLMs)~\cite{radford2021learning, li2019visualbert, lu2019vilbert,li2022blip} achieve cross-modal alignment through diverse pre-training objectives on large-scale image-text pairs~\cite{lin2014microsoft,schuhmann2022laion}. Recent advances in Large Language Models (LLMs)~\cite{touvron2023llama, touvron2023llama2} have inspired new VLM architectures that integrate LLM capabilities via MLP projection layers~\cite{liu2024llavanext,liu2024visual} or specialized modules like Q-Former~\cite{li2023blip}. Through techniques like instruction tuning~\cite{liu2024visual}, current VLMs~\cite{liu2024llavanext,liu2024visual,chen2024internvl} demonstrate enhanced performance across vision-centric tasks, including visual grounding, question answering, and image captioning, \etc. These models also enable intelligent systems for task planning and decision-making~\cite{huang2023voxposer,yokoyama2024vlfm}.

Despite their impressive capabilities~\cite{radford2021learning,liu2025mmbench,zhang2024benchmarking}, concerns remain about the robustness and generalization of VLMs, especially in safety-critical environments.  Recent studies~\cite{tong2024eyes} highlight that VLMs still exhibit deficiencies in processing visual details, such as orientation, quantity, and colour, which may be attributed to their visual representations. Moreover, VLMs remain susceptible to $L_p$-norm adversarial perturbations~\cite{cui2024robustness, mao2022understanding}, with some efforts aiming to enhance their robustness through adversarial fine-tuning techniques~\cite{mao2022understanding}. In contrast to previous works, our study specifically investigates VLM robustness against 3D variations, addressing a crucial gap in understanding their reliability for dynamic real-world applications.


\subsection{Evaluation for 3D Variation Robustness}
Robustness to 3D variations remains a longstanding challenge in computer vision. Evaluating this property often involves capturing worst-case 3D transformations, \ie the adversarial poses or viewpoints, to create challenging samples for assessment. Prior research can be categorized into three paradigms (see Fig.\ref{fig:intro}): \textbf{\emph{(A) Human-guided}} methods, like ObjectNet~\cite{barbu2019objectnet} and OOD-CV~\cite{zhao2022ood}, manually collect samples with diverse poses and viewpoints, but they lack systematic coverage of worst-case examples and are costly. \textbf{\emph{(B) Simulator-based}}~\cite{alcorn2019strike,hamdi2020towards,madan2022and} methods optimize worst-case 3D parameters via differentiable rendering of synthetic objects but require manually designed 3D assets and produce samples with limited realism. \textbf{\emph{(C) Multi-view-based}} methods, like ViewFool~\cite{dong2022viewfool} and GMVFool~\cite{ruan2023towards, ruan2023improving}, use NeRF-based~\cite{mildenhall2020nerf} representations to optimize adversarial viewpoints, but they require dense observations and struggle with background integration. In contrast, the proposed AdvDreamer employs a \textbf{\emph{(D) Generative-based}} approach, utilizing robust 3D priors from generative models. Given a single natural image, AdvDreamer generate realistic Adv-3DT samples in a zero-shot manner, providing a more efficient solution for evaluating 3D variation robustness.

\section{Methodology}
The overview of AdvDreamer is shown in Fig.~\ref{fig:framework}. 
It consists \ding{182} Zero-shot Monocular Pose Manipulation pipeline, leveraging generative 3D representations to obtain specified 3D variation samples from single-view images; \ding{183} Naturalness Reward Model, providing naturalness regularization during adversarial optimization; and \ding{184} Inverse Semantic Probability loss, ensuring the Adv-3DT samples effectively threaten VLMs and generalize across different architectures and tasks. The following sections will firstly formalize the optimization problem of AdvDreamer (Sec.\ref{sec:Problem Formulation}),  then sequentially describe the key components (Sec.\ref{sec:Framework Design}), and finally outline the full optimization algorithm (Sec.\ref{sec:Black-box}).
\subsection{Problem Formulation} \label{sec:Problem Formulation}
\noindent\textbf{Parametrization of Real-World 3D Variations.} In this paper, we primarily focus on rigid 3D transformations of foreground objects, \ie, the rotation, translation, and scaling, since they reflect the most typical 3D changes observed in real-world environments. Formally, we define a 6-dimensional vector $\boldsymbol{\Theta}=\{\alpha,\beta,\gamma,\Delta_x, \Delta_y, s\}$ to uniquely parameterize any arbitrary transformation, where $\{\alpha,\beta,\gamma\}$ denote the Tait-Bryan angles (yaw, pitch, roll) in the $Z$-$Y$-$X$ sequence, $\{\Delta_x,\Delta_y\}$ represent the translation along the $x$ and $y$ axes in the $xy$-plane, and $s$ is the uniform scaling factor. To prevent the loss of critical visual cues and ensure human recognizability, we follow previous work~\cite{dong2022viewfool} to constrain $\boldsymbol{\Theta}$ within a bounded range $[\boldsymbol{\Theta}{\min}, \boldsymbol{\Theta}{\max}]$.




\noindent\textbf{Optimization Problem.} The objective of AdvDreamer is to find an optimal distribution $p^*(\boldsymbol{\Theta})$, ensuring that Adv-3DT sampled from this distribution are both aggressive against VLMs and maintaining visual quality, which can be abstractly formulated as the following optimization problem:
\begin{equation}
\begin{gathered}
\small
    p^*(\boldsymbol{\Theta}) = \arg\max_{p(\boldsymbol{\Theta})}\mathbb{E}_{\boldsymbol{\Theta}\sim p(\boldsymbol{\Theta)}}[\mathcal{L}_{ISP}(X') + \mathcal{L}_{Nat}(X')],  \\
    \text{where}~ X' = \mathcal{T}(\boldsymbol{\Theta},X),
\label{problem1}
\end{gathered}
\end{equation}
where $\mathcal{T}(\boldsymbol{\Theta},X)$ is a transformation function to produce a novel image with the specified transformation $\boldsymbol{\Theta}$ applied to the original image $X\in\mathbb{R}^3$. To implement $\mathcal{T}(\cdot)$, we design the MPM pipeline, which will be detailed in Sec.\ref{sec:MPM}. The $\mathcal{L}_{Nat}$ and $\mathcal{L}_{ISP}$ measure the naturalness and adversarial efficacy of the transformed samples, respectively. These will be further elaborated in Sec.\ref{sec:Naturalness} and Sec.\ref{sec:Inverse Semantic}.

Optimizing over the distribution $p(\boldsymbol{\Theta})$ rather than a deterministic $\boldsymbol{\Theta}$ offers two key advantages. \emph{\textbf{1)}} As highlighted in~\cite{dong2022viewfool, ruan2023towards}, learning the underlying distribution enables comprehensive exploration of the 3D variation space, facilitating the identification of the model's vulnerable regions. \emph{\textbf{2)}} Given the stochastic nature of the generation, optimizing over a continuous distribution can mitigate the impact of appearance inconsistency, as appearance variations induced by the generation are unlikely to consistently deceive the model without optimizing the adversarial 3D variations~\cite{athalye2018synthesizing}.

\subsection{AdvDreamer Framework} \label{sec:Framework Design}
\subsubsection{Zero-shot Monocular Pose Manipulation} \label{sec:MPM}

To implement the transformation function $\mathcal{T}$, previous approaches typically rely on explicit 3D structures~\cite{alcorn2019strike,hamdi2020towards,madan2022and} or neural radiance representations~\cite{dong2022viewfool,ruan2023towards}. However, these approaches cannot accommodate real-world scenarios with limited prior knowledge. To address this, we design a simple yet effective zero-shot monocular pose manipulation (MPM) pipeline that consists of the following process:

\noindent\ding{182}~\textbf{Foreground-Background Pairs Preparation.} MPM first decompose the input image $X$ into foreground-background pairs $\{X_f, X_b\}$. Specifically, we leverage Grounded-SAM~\cite{ren2024grounded} to semantically segment the major instance as foreground $X_f$, followed by diffusion-based inpainting~\cite{Rombach_2022_CVPR,couairon2023diffedit} to obtain a complete background $X_b$. This process is denoted as $\{X_f,X_b\}\!\!=\!\!\mathcal{F}(X)$. Additionally, we support directly generating synthetic $\{X_f, X_b\}$ pairs through Stable-Diffusion~\cite{Rombach_2022_CVPR} to enhance sample diversity.


\noindent\ding{183}~\textbf{Object-level Pose Manipulation.} For each optimization iteration, we transform the foreground $X_f$ into a batch of $X'_f$ under $\boldsymbol{\Theta}$ sampled from current $p(\boldsymbol{\Theta})$. In this stage, MPM leverages the robust priors provided by a pre-trained Large Reconstruction Model (\eg, TripoSR~\cite{TripoSR2024}) to construct single-view-based 3D representations and apply sampled transformations. This process is formulated as $X_f' = \mathcal{R}_{\textbf{w}_0}(X_f, \boldsymbol{\Theta})$, where $\boldsymbol{\Theta} \sim p(\boldsymbol{\Theta})$ and $\mathcal{R}_\textbf{w}(\cdot)$ denotes nerual rendering function with weights $\textbf{w}_0$.

\noindent\ding{184}~\textbf{Re-Composition.} We then compose the transformed foreground $X'_f$ with background $X_b$ through a pre-trained diffusion-based composition model~\cite{chen2024anydoor, zhang2025scaling}, to handles shadows and occlusions. This process is defined as $X'=\mathcal{C}_{\textbf{w}_1}(X_f', X_b)$, where $\textbf{w}_1$ denotes diffusion model's weights. 

Combining the above stages, the transformation process by MPM can be formally represented as follows:
\begin{equation} \label{generate}
\begin{gathered}
    X' = \mathcal{T}(\boldsymbol{\Theta}, X)=\mathcal{C}_{\textbf{w}_1}(\mathcal{R}_{\textbf{w}_0} (X_f, \boldsymbol{\Theta}),X_b), \\
   \text{where} ~~~\{X_f, X_b\} = \mathcal{F}(X).
\end{gathered}
\end{equation}

\subsubsection{Naturalness Reward Model} 
Despite the MPM pipeline utilizing generative priors to achieve realistic 3D variation samples, we observe that directly applying adversarial optimization frequently leads to unnatural artifacts, such as shape distortions and physically implausible cases in Fig.~\ref{fig:failure-sample}. These issues stem from two factors: generalization limitations of existing generative models and the inherent high semantic bias in low-quality samples. To address this challenge, we propose a novel Naturalness Reward Model (NRM) that continuously regularizes $p(\boldsymbol{\Theta})$ during optimization based on sample naturalness, preventing convergence to low-quality pseudo-optimal regions.

\begin{figure}[t]
  \centering
  \includegraphics[height=2.2cm]{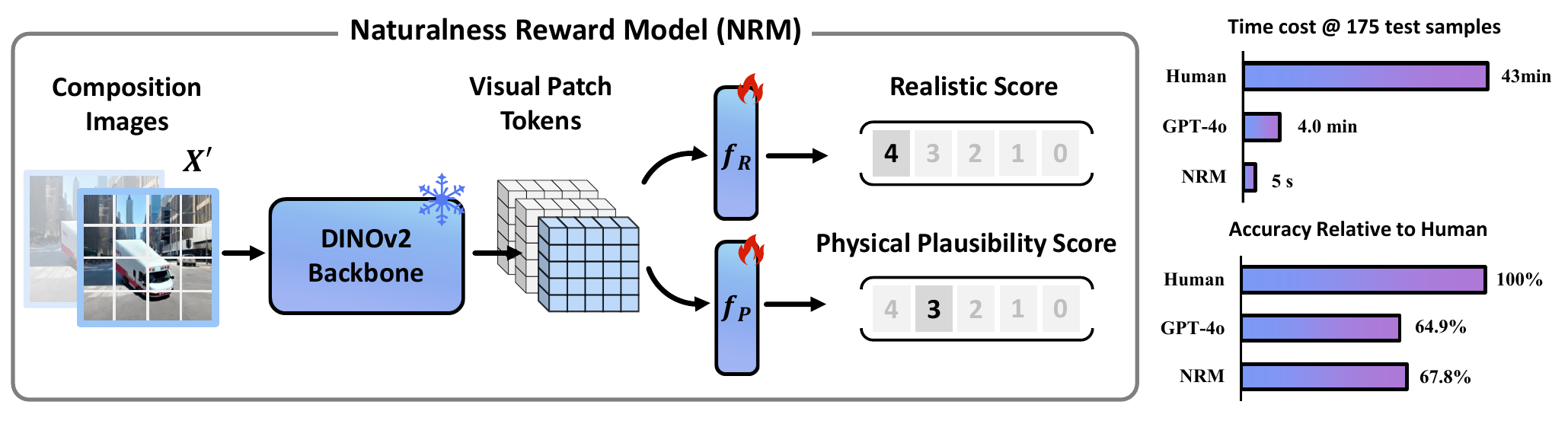}
  \vspace{-0.2cm}
  \caption{Architecture of NRM and computational cost and accuracy of naturalness assessment across Human, GPT, and NRM.}
  \vspace{-0.5cm}
  \label{fig:NRM}
\end{figure}

As shown in Fig.\ref{fig:NRM}, the NRM employs DINOv2\cite{oquab2023dinov2} as its backbone, chosen for several reasons. \textbf{\emph{Firstly}}, DINO captures finer visual contextual features compared to traditional pre-trained vision encoders~\cite{tong2024eyes}, aiding better convergence during training. \textbf{\emph{Secondly}}, unlike vision-language aligned encoders such as CLIP, DINO’s visual features are language-agnostic, making them more faithfully reflect image quality rather than semantic consistency. We use DINOv2 to extract image tokens from the input Adv-3DT samples, followed by dual-stream prediction heads to estimate visual fidelity and physical plausibility scores: 
\begin{equation}
\text{Score}_R(X') = f_{R}(\mathcal{E}_{D}(X')),~\text{Score}_P(X') = f_{P}(\mathcal{E}_{D}(X')),
\end{equation}
where $f_{R}(\cdot)$ and $f_{P}(\cdot)$ are two-layer MLPs. To train the NRM, we construct a large-scale naturalness preference dataset with fine-grained scoring. We generate numerous samples from the MPM pipeline on ImageNet and synthetic images, scoring each on a 5-point scale for visual fidelity and physical plausibility. These scores are initially annotated automatically by GPT-4o with Chain-of-Thought prompts and verified by volunteers for correction. This process resulted in 120K high-quality training samples.

\label{sec:Naturalness}
The architecture and training parameters of NRM are detailed in Appendix \textcolor{iccvblue}{B}. Our experimental results demonstrate that the naturalness feedback provided by NRM highly aligns with human evaluation. In terms of prediction efficiency and stability, it outperforms methods that directly employ GPT-4o for naturalness assessment (see Fig.~\ref{fig:NRM} and Appendix \textcolor{iccvblue}{C.2}). Based on the NRM's predictions, we define the naturalness regularization loss for the optimization as:
\begin{equation}
\mathcal{L}_{Nat}(X') = -\frac{1}{2}(\text{Score}_R(X') + \text{Score}_P(X'))
\end{equation}


\subsubsection{Inverse Semantic Probability Loss} 
\label{sec:Inverse Semantic}
Previous studies typically employ task-specific optimization objectives with end-to-end manner to obtain worst-case Adv-3DT samples. However, directly applying this approach to VLMs with substantial parameters and diverse architectures spanning various tasks is challenging. To address this, we design the Inverse Semantic Probability (ISP) loss, which minimizes the probability that VLMs assign to the correct semantic attributes of Adv-3DT samples.

This design have two key considerations. \emph{\textbf{1)}} It is based solely on the visual and text encoder $\mathcal{E}_v$, $\mathcal{E}_t$, which are fundamental components across modern VLMs, ensuring architecture-agnostic applicability. \emph{\textbf{2)}} By operating in the image-text alignment space rather than model-specific layers or task-specific heads, it enables task generalization. 

Given a generated sample $X'$ and a semantic label set $\mathcal{Y}=\{y_i\}_{i=1}^N$ containing the ground-truth label $y_t$, we first compute a semantic similarity vector $\mathbf{S}$:
\begin{equation}
\begin{gathered}
\mathbf{S}=\{s_i\}_{i=1}^N,~
\text{where}~~s_i = \cos \left[ \mathcal{E}_v(X'),\mathcal{E}_t(\mathcal{P}(y_i)) \right],
\end{gathered}
\end{equation}
where $\cos(\cdot)$ computes cosine similarity and $\mathcal{P}(\cdot)$ follows the template \textit{"a photo of a \{$y_i$\}"} in~\cite{radford2021learning}. The vector $\mathbf{S}$ captures the representation similarities between $X'$ and each semantic label. We then derive the conditional semantic probability through softmax transformation:
\begin{equation}
p(y_i|X') = \exp(s_i) / \textstyle \sum_{j=1}^{N} \exp(s_j).
\end{equation}
Then, the $\mathcal{L}_\text{ISP}$ is defined as the negative log-likelihood of the ground-truth's semantic probability:
\begin{equation}
\mathcal{L}_{\text{ISP}}(X',\mathcal{Y}) = -\log p(y_t|X').
\end{equation}
Intuitively, $\mathcal{L}_{\text{ISP}}$ quantifies the likelihood of $X'$ being misclassified by comparing its similarity to the ground-truth and irrelevant categories. Maximizing it directs the $p(\boldsymbol{\Theta})$ towards regions that effectively deceive the model.

\begin{table*}[t]
\begin{minipage}[t]{0.48\textwidth}
\caption{Zero-shot Classification Accuracy (\%) and degradation (\textcolor{Maroon}{$\downarrow$\%}) Under Adv-3DT Samples Generated from ImageNet~\cite{deng2009imagenet}.}
\vspace{-0.2cm}
\setlength\tabcolsep{3.5pt}
\centering
\scalebox{0.75}{
\begin{tabular}{lc|cccc}
\hline
\multirow{2}{*}{Target Models} & \multirow{2}{*}{\#Params} & \multicolumn{4}{c}{Methods}                                                     \\ \cline{3-6} 
                             &                       & Clean & Random   & $p^*(\boldsymbol{\Theta})$ & $\boldsymbol{\Theta}^*$ \\ \hline
OpenCLIP~\footnotesize{ViT-B/16}~\cite{openclip}          &  149M                          & 98.0  & 62.6~\footnotesize{$(\textcolor{Maroon}{\downarrow\! 35})$}  & 54.0~\footnotesize{$(\textcolor{Maroon}{\downarrow\! 44})$} & \textbf{18.0}~\footnotesize{$\boldsymbol{(\textcolor{Maroon}{\downarrow\! 80})}$} \\
OpenCLIP~\footnotesize{ViT-L/14}~\cite{openclip}           &  428M                         & 94.4  & 61.7~\footnotesize{$(\textcolor{Maroon}{\downarrow\! 33})$}  & 50.9~\footnotesize{$(\textcolor{Maroon}{\downarrow\! 44})$} & \textbf{15.3}~\footnotesize{$\boldsymbol{(\textcolor{Maroon}{\downarrow\! 79})}$} \\
OpenCLIP~\footnotesize{ViT-G/14}~\cite{openclip}           & 2.5B                          & 96.4  & 63.5~\footnotesize{$(\textcolor{Maroon}{\downarrow\! 33})$}  & 53.5~\footnotesize{$(\textcolor{Maroon}{\downarrow\! 43})$} & \textbf{18.7}~\footnotesize{$\boldsymbol{(\textcolor{Maroon}{\downarrow\! 78})}$} \\ \hline
BLIP~\footnotesize{ViT-B/16}~\cite{li2022blip}              & 583M                          & 83.0  & 56.0~\footnotesize{$(\textcolor{Maroon}{\downarrow\! 27})$}  & 51.3~\footnotesize{$(\textcolor{Maroon}{\downarrow\! 32})$} & \textbf{17.3}~\footnotesize{$\boldsymbol{(\textcolor{Maroon}{\downarrow\! 66})}$} \\
BLIP-2~\footnotesize{ViT-L/14}~\cite{li2023blip}            & 3.4B                          & 84.0  & 57.0~\footnotesize{$(\textcolor{Maroon}{\downarrow\! 27})$}  & 52.2~\footnotesize{$(\textcolor{Maroon}{\downarrow\! 32})$} & \textbf{18.7}~\footnotesize{$\boldsymbol{(\textcolor{Maroon}{\downarrow\! 65})}$} \\ 
BLIP-2~\footnotesize{ViT-G/14}~\cite{li2023blip}           &  4.1B                         & 81.0  & 55.7~\footnotesize{$(\textcolor{Maroon}{\downarrow\! 25})$}  & 49.1~\footnotesize{$(\textcolor{Maroon}{\downarrow\! 32})$} & \textbf{15.7}~\footnotesize{$\boldsymbol{(\textcolor{Maroon}{\downarrow\! 65})}$} \\ \hline
\end{tabular}}
\label{exp:1}
\end{minipage}\hfill
\begin{minipage}[t]{0.48\textwidth}
\renewcommand\arraystretch{1.03}
\caption{Cross-model transferability of Adv-3DT samples, utilizing OpenCLIP and BLIP as surrogate models.}
\vspace{-0.2cm}
\setlength\tabcolsep{5.0pt}
\centering
\scalebox{0.65}{
\begin{tabular}{l|c|cc|cc}
\hline
\multirow{2}{*}{Target Model} & \multirow{2}{*}{Clean} & \multicolumn{2}{c|}{OpenCLIP~\cite{openclip}}                        & \multicolumn{2}{c}{BLIP~\cite{li2022blip}}                             \\ \cline{3-6} 
                              &                        & $p^*(\boldsymbol{\Theta})$ & $\boldsymbol{\Theta}^*$ & $p^*(\boldsymbol{\Theta})$ & $\boldsymbol{\Theta}^*$ \\ \hline
ALBEF~\cite{li2021align}                         & 65.0                   & \textbf{37.2}~\footnotesize{$(\textcolor{Maroon}{\downarrow\! 28})$}                       & \underline{27.7}~\footnotesize{$(\textcolor{Maroon}{\downarrow\! 37})$}                    & \textbf{40.2}~\footnotesize{$(\textcolor{Maroon}{\downarrow\! 25})$}                       & \underline{25.7}~\footnotesize{$(\textcolor{Maroon}{\downarrow\! 39})$}                    \\
OpenAI CLIP~\footnotesize{ViT-B/16}~\cite{radford2021learning}        & 85.3                   & 48.7~\footnotesize{$(\textcolor{Maroon}{\downarrow\! 37})$}                       & 29.7~\footnotesize{$(\textcolor{Maroon}{\downarrow\! 56})$}                    & 52.8~\footnotesize{$(\textcolor{Maroon}{\downarrow\! 32})$}                       & 31.0~\footnotesize{$(\textcolor{Maroon}{\downarrow\! 54})$}                    \\
OpenCLIP~\footnotesize{ViT-B/16}~\cite{openclip}          & 97.7                   & \cellcolor{gray!25}54.0~\footnotesize{$\boldsymbol{(\textcolor{Maroon}{\downarrow\! 44})}$}                       & \cellcolor{gray!25}\textbf{18.0}~\footnotesize{$\boldsymbol{(\textcolor{Maroon}{\downarrow\! 80})}$}                    & 59.3~\footnotesize{$\boldsymbol{(\textcolor{Maroon}{\downarrow\! 38})}$}                       & 34.7~\footnotesize{$(\textcolor{Maroon}{\downarrow\! \underline{63}})$}                    \\
Meta-CLIP~\footnotesize{ViT-B/16}~\cite{xudemystifying}          & 91.0                   & 49.3~\footnotesize{$(\textcolor{Maroon}{\downarrow\! 42})$}                       & 30.0~\footnotesize{$(\textcolor{Maroon}{\downarrow\! \underline{61}})$}                    & 53.8~\footnotesize{$(\textcolor{Maroon}{\downarrow\! \underline{37}})$}                       & 29.7~\footnotesize{$(\textcolor{Maroon}{\downarrow\! 61})$}                    \\
BLIP~\footnotesize{ViT-B/16}~\cite{li2022blip}               & 82.7                   & \underline{48.4}~\footnotesize{$(\textcolor{Maroon}{\downarrow\! 34})$}                       & 30.7~\footnotesize{$(\textcolor{Maroon}{\downarrow\! 52})$}                    & \cellcolor{gray!25}\underline{51.3}~\footnotesize{$(\textcolor{Maroon}{\downarrow\! 31})$}                       & \cellcolor{gray!25}\textbf{17.3}~\footnotesize{$\boldsymbol{(\textcolor{Maroon}{\downarrow\! 65})}$}                    \\
BLIP-2~\footnotesize{ViT-L/14}~\cite{li2023blip}             & 84.0                   & 50.0~\footnotesize{$(\textcolor{Maroon}{\downarrow\! 34})$}                       & 32.0~\footnotesize{$(\textcolor{Maroon}{\downarrow\! 52})$}                    & 52.3~\footnotesize{$(\textcolor{Maroon}{\downarrow\! 32})$}                       & 27.7~\footnotesize{$(\textcolor{Maroon}{\downarrow\! 56})$}                    \\
SigLIP~\footnotesize{ViT-B/16}~\cite{zhai2023sigmoid}             & 95.3                   & 55.3~\footnotesize{${(\textcolor{Maroon}{\downarrow\! \underline{40}})}$}                     & 34.0~\footnotesize{$(\textcolor{Maroon}{\downarrow\! \underline{61}})$}                    & 59.3~\footnotesize{$(\textcolor{Maroon}{\downarrow\! 36})$}                       & 34.7~\footnotesize{$(\textcolor{Maroon}{\downarrow\! 61})$}                    \\ \hline
\end{tabular}}
\label{exp:2}
\end{minipage}
\vspace{-0.5cm}
\end{table*}

\subsection{Query-Based Adversarial Optimization} \label{sec:Black-box}
To solve Problem~\eqref{problem1}, we parameterize the adversarial distribution $p(\boldsymbol{\Theta})$ as a \emph{\textbf{Multivariate Gaussian}} form:
\begin{equation}
\boldsymbol{\Theta} = \mathbf{A}\cdot\tanh(\mathbf{z})+\mathbf{B},~\text{where}~\mathbf{z}\sim \mathcal{N}(\boldsymbol{\mu},\boldsymbol{\Sigma}),
\end{equation}
where $\mathbf{A}\!\!=\!\!(\boldsymbol{\Theta}_{\min}\!-\!\boldsymbol{\Theta}_{\max})/2$, $\mathbf{A}\!\!=\!\!(\boldsymbol{\Theta}_{\min}\!+\!\boldsymbol{\Theta}_{\max})/2$, $\mathbf{z}$ follows a Gaussian Distribution with mean $\boldsymbol{\mu}\!\in\!\mathbb{R}^6$ and covariance $\boldsymbol{\Sigma}\!\in\!\mathbb{R}^{6\times6}$. Thus, we rewrite \eqref{problem1} in a specific form:
\begin{equation}
\begin{gathered}
\arg \max_{\boldsymbol{\mu},\boldsymbol{\Sigma}}\mathbb{E}_{\mathbf{z}\sim\mathcal{N}(\boldsymbol{\mu},\boldsymbol{\Sigma})}[\mathcal{L}_{\text{ISP}}(X',\mathcal{Y})+\mathcal{L}_{\text{Nat}}(X')],\\
\text{where}~X'=\mathcal{T}(\mathbf{A}\cdot\tanh(\mathbf{z})+\mathbf{B},X).
\end{gathered}
\end{equation}
Solving this problem typically involves computing the gradient of the loss \wrt $\boldsymbol{\mu}$, $\boldsymbol{\Sigma}$ and updating them via gradient ascent. However, the forward steps involve multiple components, \eg, the denoise process, introducing uncertainty in the gradient propagation path, making gradient-based optimization challenging~\cite{song2020denoising}. Therefore, we adopt the Covariance Matrix Adaptation Evolution Strategy~\cite{hansen2016cma,golovin2017google}, an efficient query-based black-box optimizer. At each iteration $t$, we: \textbf{\emph{1)}} Sample $K$ candidates $\{\mathbf{z}^{t+1}_i\}_{i=1}^K$ from $\mathcal{N}(\boldsymbol{\mu}^t,\boldsymbol{\Sigma}^t)$. \textbf{\emph{2)}} Generate corresponding samples $\{(X')^{t+1}_i\}_{i=1}^K$. \textbf{\emph{3)}} Update distribution parameters $\boldsymbol{\mu}^t,\boldsymbol{\Sigma}^t$ as follows:
\vspace{-0.2cm}
\begin{equation} \label{update mu}
\boldsymbol{\mu}^{t+1} = \sum_{i=1}^{k}w_i\cdot \mathbf{z}_{(i:k)}^{t+1},~~\text{where}~\sum_{i=1}^{k}w_i = 1,
\end{equation}
\begin{equation}
\begin{gathered} \label{sigma}
\boldsymbol{\Sigma}^{t+1} = (1-\eta_1-\eta_{\mu})\cdot\boldsymbol{\Sigma}^t+\eta_1\cdot p^{t+1}_{\Sigma}(p^{t+1}_{\Sigma})^T+ \\
\eta_{\mu}\cdot\sum_{i=1}^{h}w_i\cdot(\frac{\mathbf{z}_{(i:k)}^{t+1}-\boldsymbol{\mu}^{t}}{\boldsymbol{\sigma}^t})(\frac{\mathbf{z}_{(i:k)}^{t+1}-\boldsymbol{\mu}^{t}}{\boldsymbol{\sigma}^t})^T,
\end{gathered}
\end{equation}
here, $\mathbf{z}_{(i:k)}$ denotes samples ranked by losses: we first select top-10 samples based on $\mathcal{L}_{\text{ISP}}$, then choose the top-k~(k=5) with highest $\mathcal{L}_{\text{Nat}}$ for updates. $w_i=1/k$ are importance weights, $\eta_1$ and $\eta_{\mu}$ are learning rates, $p_{\Sigma}$ tracks the evolution path, and $\boldsymbol{\sigma}^t$ is the step size updated via Cumulative Step-size Adaptation (CSA)~\cite{hansen2016cma}. Due to space constraints, complete derivations of Eq.~\eqref{update mu}, \eqref{sigma} and hyperparameter settings are provided in the Appendix.\textcolor{iccvblue}{A.1}. We also provide the pseudocode for optimization process in Appendix~\textcolor{iccvblue}{A.2}.

\vspace{-0.1cm}

\section{Experiments}
\begin{table*}[t]
\caption{\textbf{Performance Degradation (\textcolor{Maroon}{$\downarrow$}) of VLMs on Image Captioning and VQA Tasks.} For closed-source VLMs (GPT-4o/4o-mini), we performed transfer-based attacks, using OpenAI-CLIP (ViT-L@336px) as surrogate encoder to optimize $p^*(\boldsymbol{\Theta})$. For other VLMs, $p^*(\boldsymbol{\Theta})$ are optimized \wrt respective image encoders (\eg, Eva-CLIP and OpenAI CLIP, \etc.). Metrics: C-CIDEr, S-SPICE, B-BLUE@4.}
\vspace{-0.1cm}
\setlength\tabcolsep{9.5pt}
\renewcommand\arraystretch{0.9}
\centering
\scalebox{0.55}{
\begin{tabular}{c|c|cccc|cccc|c}
\hline
                                      &                            & \multicolumn{4}{c|}{COCO Caption~\cite{chen2015microsoft}}                                                                                             & \multicolumn{4}{c|}{NoCaps~\cite{agrawal2019nocaps}}                                                                                                  & VQA-V2~\cite{goyal2017making}                       \\ \cline{3-11} 
\multirow{-2}{*}{Model}               & \multirow{-2}{*}{Method}   & C                            & S                            & {B@4} & GPT-Score                    & C                            & S                           & {B@4} & GPT-Score                    & GPT-Acc                      \\ \hline
                                      & Clean                      & 133.4                        & 22.8                         & 36.0                             & 24.7                         & 98.1                         & 14.1                        & 44.1                             & 24.4                         & 53.4                         \\
                                      & Random                     & 91.8 $(\textcolor{orange}{\downarrow\! 41.6})$  & 17.6 $(\textcolor{orange}{\downarrow\! 5.2})$  & 23.5 $(\textcolor{orange}{\downarrow\! 12.5})$  & 19.5 $(\textcolor{orange}{\downarrow\! 5.2})$  & 60.2 $(\textcolor{orange}{\downarrow\! 37.9})$  & 10.1 $(\textcolor{orange}{\downarrow\! 4.0})$  & 28.6 $(\textcolor{orange}{\downarrow\! 15.5})$  & 20.9 $(\textcolor{orange}{\downarrow\! 3.5})$  & 34.0 $(\textcolor{orange}{\downarrow\! 19.4})$  \\
                                      & $p^*(\boldsymbol{\Theta})$ & 83.8 $(\textcolor{Maroon}{\downarrow\! 49.6})$  & 16.7 $(\textcolor{Maroon}{\downarrow\! 6.1})$  & 23.3 $(\textcolor{Maroon}{\downarrow\! 12.7})$  & 18.9 $(\textcolor{Maroon}{\downarrow\! 5.8})$  & 55.7 $(\textcolor{Maroon}{\downarrow\! 42.4})$  & 9.3 $(\textcolor{Maroon}{\downarrow\! 4.8})$  & 26.9 $(\textcolor{Maroon}{\downarrow\! 17.2})$  & 17.1 $(\textcolor{Maroon}{\downarrow\! 7.3})$  & 31.7 $(\textcolor{Maroon}{\downarrow\! 21.7})$  \\
\multirow{-4}{*}{BLIP-2~\footnotesize{FlanT5XL} (3.4B)~\cite{li2022blip}}     & \cellcolor[gray]{0.85}$\boldsymbol{\Theta}^*$    & \cellcolor[gray]{0.85}$\boldsymbol{72.2}$ $\boldsymbol{(\textcolor{Maroon}{\downarrow\! 61.2})}$  & \cellcolor[gray]{0.85}$\boldsymbol{14.7}$ $\boldsymbol{(\textcolor{Maroon}{\downarrow\! 8.1})}$  & \cellcolor[gray]{0.85}$\boldsymbol{21.5}$ $\boldsymbol{(\textcolor{Maroon}{\downarrow\! 14.5})}$  & \cellcolor[gray]{0.85}$\boldsymbol{17.0}$ $\boldsymbol{(\textcolor{Maroon}{\downarrow\! 7.7})}$  & \cellcolor[gray]{0.85}$\boldsymbol{40.0}$ $\boldsymbol{(\textcolor{Maroon}{\downarrow\! 58.1})}$  & \cellcolor[gray]{0.85}$\boldsymbol{7.3}$ $\boldsymbol{(\textcolor{Maroon}{\downarrow\! 6.8})}$  & \cellcolor[gray]{0.85}$\boldsymbol{22.4}$ $\boldsymbol{(\textcolor{Maroon}{\downarrow\! 21.7})}$  & \cellcolor[gray]{0.85}$\boldsymbol{15.0}$ $\boldsymbol{(\textcolor{Maroon}{\downarrow\! 9.4})}$  & \cellcolor[gray]{0.85}$\boldsymbol{28.0}$ $\boldsymbol{(\textcolor{Maroon}{\downarrow\! 25.4})}$  \\ \hline
                                      & Clean                      & 126.1                        & 25.7                         & 30.4                             & 25.0                         & 118.7                        & 16.7                        & 50.6                             & 24.7                         & 68.6                         \\
                                      & Random                     & 93.4 $(\textcolor{orange}{\downarrow\! 32.7})$  & 18.6 $(\textcolor{orange}{\downarrow\! 7.1})$  & 25.6 $(\textcolor{orange}{\downarrow\! 4.8})$  & 20.0 $(\textcolor{orange}{\downarrow\! 5.0})$  & 75.3 $(\textcolor{orange}{\downarrow\! 43.4})$  & 11.3 $(\textcolor{orange}{\downarrow\! 5.4})$  & 35.7 $(\textcolor{orange}{\downarrow\! 14.9})$  & 21.7 $(\textcolor{orange}{\downarrow\! 3.0})$  & 56.3 $(\textcolor{orange}{\downarrow\! 12.3})$  \\
                                      & $p^*(\boldsymbol{\Theta})$ & 85.8 $(\textcolor{Maroon}{\downarrow\! 40.3})$  & 16.7 $(\textcolor{Maroon}{\downarrow\! 9.0})$  & 22.9 $(\textcolor{Maroon}{\downarrow\! 7.5})$  & 18.9 $(\textcolor{Maroon}{\downarrow\! 6.1})$  & 60.0 $(\textcolor{Maroon}{\downarrow\! 58.7})$  & 9.7 $(\textcolor{Maroon}{\downarrow\! 7.0})$  & 29.6 $(\textcolor{Maroon}{\downarrow\! 21.0})$  & 17.4 $(\textcolor{Maroon}{\downarrow\! 7.3})$  & 54.3 $(\textcolor{Maroon}{\downarrow\! 14.3})$  \\
\multirow{-4}{*}{LLaVa-1.5~\footnotesize{Vicuna} (7B)~\cite{liu2024visual}}   & \cellcolor[gray]{0.85}$\boldsymbol{\Theta}^*$    & \cellcolor[gray]{0.85}$\boldsymbol{73.2}$ $\boldsymbol{(\textcolor{Maroon}{\downarrow\! 52.9})}$  & \cellcolor[gray]{0.85}$\boldsymbol{14.4}$ $\boldsymbol{(\textcolor{Maroon}{\downarrow\! 11.3})}$  & \cellcolor[gray]{0.85}$\boldsymbol{20.5}$ $\boldsymbol{(\textcolor{Maroon}{\downarrow\! 9.9})}$  & \cellcolor[gray]{0.85}$\boldsymbol{16.5}$ $\boldsymbol{(\textcolor{Maroon}{\downarrow\! 8.5})}$  & \cellcolor[gray]{0.85}$\boldsymbol{45.8}$ $\boldsymbol{(\textcolor{Maroon}{\downarrow\! 72.9})}$  & \cellcolor[gray]{0.85}$\boldsymbol{8.2}$ $\boldsymbol{(\textcolor{Maroon}{\downarrow\! 8.5})}$  & \cellcolor[gray]{0.85}$\boldsymbol{25.5}$ $\boldsymbol{(\textcolor{Maroon}{\downarrow\! 25.1})}$  & \cellcolor[gray]{0.85}$\boldsymbol{15.6}$ $\boldsymbol{(\textcolor{Maroon}{\downarrow\! 9.1})}$  & \cellcolor[gray]{0.85}$\boldsymbol{53.8}$ $\boldsymbol{(\textcolor{Maroon}{\downarrow\! 14.8})}$  \\ \hline
                                      & Clean                      & 41.6 & 11.2 & 6.9      & 22.9 & 30.5 & 6.9 & 10.8     & 22.6 & 71.2 \\
                                      & Random                     & 25.1 $(\textcolor{orange}{\downarrow\! 16.5})$ & 6.9 $(\textcolor{orange}{\downarrow\! 4.3})$  & 6.0 $(\textcolor{orange}{\downarrow\! 0.9})$      & 17.5 $(\textcolor{orange}{\downarrow\! 5.4})$ & 18.3 $(\textcolor{orange}{\downarrow\! 12.2})$ & 4.7 $(\textcolor{orange}{\downarrow\! 2.2})$ & 7.1 $(\textcolor{orange}{\downarrow\! 3.7})$      & 21.0 $(\textcolor{orange}{\downarrow\! 1.6})$ & 54.3 $(\textcolor{orange}{\downarrow\! 16.9})$ \\
                                      & $p^*(\boldsymbol{\Theta})$ & 21.9 $(\textcolor{Maroon}{\downarrow\! 19.7})$ & 6.5 $(\textcolor{Maroon}{\downarrow\! 4.7})$  & 5.3 $(\textcolor{Maroon}{\downarrow\! 1.6})$      & 16.9 $(\textcolor{Maroon}{\downarrow\! 6.0})$ & 13.6 $(\textcolor{Maroon}{\downarrow\! 16.9})$ & 4.0 $(\textcolor{Maroon}{\downarrow\! 2.9})$ & 6.5 $(\textcolor{Maroon}{\downarrow\! 4.3})$      & 15.7 $(\textcolor{Maroon}{\downarrow\! 6.9})$ & 53.0 $(\textcolor{Maroon}{\downarrow\! 18.2})$ \\
\multirow{-4}{*}{LLaVa-1.6~\footnotesize{Vicuna} (13B)~\cite{liu2024llavanext}}   & \cellcolor[gray]{0.85}$\boldsymbol{\Theta}^*$    & \cellcolor[gray]{0.85}$\boldsymbol{16.7}$ $\boldsymbol{(\textcolor{Maroon}{\downarrow\! 24.9})}$ & \cellcolor[gray]{0.85}$\boldsymbol{5.5}$ $\boldsymbol{(\textcolor{Maroon}{\downarrow\! 5.7})}$  & \cellcolor[gray]{0.85}$\boldsymbol{4.6}$ $\boldsymbol{(\textcolor{Maroon}{\downarrow\! 2.3})}$      & \cellcolor[gray]{0.85}$\boldsymbol{15.4}$ $\boldsymbol{(\textcolor{Maroon}{\downarrow\! 7.5})}$ & \cellcolor[gray]{0.85}$\boldsymbol{10.5}$ $\boldsymbol{(\textcolor{Maroon}{\downarrow\! 20.0})}$ & \cellcolor[gray]{0.85}$\boldsymbol{3.0}$ $\boldsymbol{(\textcolor{Maroon}{\downarrow\! 3.9})}$ & \cellcolor[gray]{0.85}$\boldsymbol{4.3}$ $\boldsymbol{(\textcolor{Maroon}{\downarrow\! 6.5})}$      & \cellcolor[gray]{0.85}$\boldsymbol{13.5}$ $\boldsymbol{(\textcolor{Maroon}{\downarrow\! 9.1})}$ & \cellcolor[gray]{0.85}$\boldsymbol{46.9}$ $\boldsymbol{(\textcolor{Maroon}{\downarrow\! 24.3})}$ \\ \hline
                                      & Clean                      & 127.0                        & 22.6                         & 34.9                             & 24.0                         & 76.4                         & 12.1                        & 29.7                             & 22.5                         & 53.9                         \\
                                      & Random                     & 71.6 $(\textcolor{orange}{\downarrow\! 55.4})$  & 15.9 $(\textcolor{orange}{\downarrow\! 6.7})$  & 22.5 $(\textcolor{orange}{\downarrow\! 12.4})$  & 21.0 $(\textcolor{orange}{\downarrow\! 3.0})$  & 55.4 $(\textcolor{orange}{\downarrow\! 21.0})$  & 8.5 $(\textcolor{orange}{\downarrow\! 3.6})$  & 23.7 $(\textcolor{orange}{\downarrow\! 6.0})$  & 21.1 $(\textcolor{orange}{\downarrow\! 1.4})$  & 49.7 $(\textcolor{orange}{\downarrow\! 4.2})$  \\
                                      & $p^*(\boldsymbol{\Theta})$ & 67.9 $(\textcolor{Maroon}{\downarrow\! 59.1})$  & 13.8 $(\textcolor{Maroon}{\downarrow\! 8.8})$  & 17.8 $(\textcolor{Maroon}{\downarrow\! 17.1})$  & 17.2 $(\textcolor{Maroon}{\downarrow\! 6.8})$  & 45.4 $(\textcolor{Maroon}{\downarrow\! 31.0})$  & 7.6 $(\textcolor{Maroon}{\downarrow\! 4.5})$  & 19.2 $(\textcolor{Maroon}{\downarrow\! 10.5})$  & 16.2 $(\textcolor{Maroon}{\downarrow\! 6.3})$  & 44.9 $(\textcolor{Maroon}{\downarrow\! 9.0})$  \\
\multirow{-4}{*}{Otter~\footnotesize{MPT} (7B)~\cite{li2023mimic}}           & \cellcolor[gray]{0.85}$\boldsymbol{\Theta}^*$    & \cellcolor[gray]{0.85}$\boldsymbol{58.9}$ $\boldsymbol{(\textcolor{Maroon}{\downarrow\! 68.1})}$  & \cellcolor[gray]{0.85}$\boldsymbol{13.0}$ $\boldsymbol{(\textcolor{Maroon}{\downarrow\! 9.6})}$  & \cellcolor[gray]{0.85}$\boldsymbol{17.5}$ $\boldsymbol{(\textcolor{Maroon}{\downarrow\! 17.4})}$  & \cellcolor[gray]{0.85}$\boldsymbol{15.8}$ $\boldsymbol{(\textcolor{Maroon}{\downarrow\! 8.2})}$  & \cellcolor[gray]{0.85}$\boldsymbol{36.0}$ $\boldsymbol{(\textcolor{Maroon}{\downarrow\! 40.4})}$  & \cellcolor[gray]{0.85}$\boldsymbol{6.1}$ $\boldsymbol{(\textcolor{Maroon}{\downarrow\! 6.0})}$  & \cellcolor[gray]{0.85}$\boldsymbol{15.7}$ $\boldsymbol{(\textcolor{Maroon}{\downarrow\! 14.0})}$  & \cellcolor[gray]{0.85}$\boldsymbol{14.6}$ $\boldsymbol{(\textcolor{Maroon}{\downarrow\! 7.9})}$  & \cellcolor[gray]{0.85}$\boldsymbol{42.0}$ $\boldsymbol{(\textcolor{Maroon}{\downarrow\! 11.9})}$  \\ \hline
                                      & Clean                      & 78.5                         & 23.3                         & 23.4                             & 24.0                         & 60.8                         & 17.2                        & 28.5                             & 24.7                         & 51.3                         \\
                                      & Random                     & 58.2 $(\textcolor{orange}{\downarrow\! 20.3})$  & 16.5 $(\textcolor{orange}{\downarrow\! 6.8})$  & 17.9 $(\textcolor{orange}{\downarrow\! 5.5})$  & 19.7 $(\textcolor{orange}{\downarrow\! 4.3})$  & 38.0 $(\textcolor{orange}{\downarrow\! 22.8})$  & 11.1 $(\textcolor{orange}{\downarrow\! 6.1})$  & 19.8 $(\textcolor{orange}{\downarrow\! 8.7})$  & 21.9 $(\textcolor{orange}{\downarrow\! 2.8})$  & 43.7 $(\textcolor{orange}{\downarrow\! 7.6})$  \\
                                      & $p^*(\boldsymbol{\Theta})$ & 52.3 $(\textcolor{Maroon}{\downarrow\! 26.2})$  & 16.4 $(\textcolor{Maroon}{\downarrow\! 6.9})$  & 16.3 $(\textcolor{Maroon}{\downarrow\! 7.1})$  & 18.1 $(\textcolor{Maroon}{\downarrow\! 5.9})$  & 32.9 $(\textcolor{Maroon}{\downarrow\! 27.9})$  & 10.2 $(\textcolor{Maroon}{\downarrow\! 7.0})$  & 18.4 $(\textcolor{Maroon}{\downarrow\! 10.1})$  & 16.7 $(\textcolor{Maroon}{\downarrow\! 8.0})$  & 41.0 $(\textcolor{Maroon}{\downarrow\! 10.3})$  \\
\multirow{-4}{*}{MiniGPT-4~\footnotesize{Llama-2} (7B)~\cite{zhu2023minigpt}} & \cellcolor[gray]{0.85}$\boldsymbol{\Theta}^*$    & \cellcolor[gray]{0.85}$\boldsymbol{42.2}$ $\boldsymbol{(\textcolor{Maroon}{\downarrow\! 36.3})}$  & \cellcolor[gray]{0.85}$\boldsymbol{14.1}$ $\boldsymbol{(\textcolor{Maroon}{\downarrow\! 9.2})}$  & \cellcolor[gray]{0.85}$\boldsymbol{12.9}$ $\boldsymbol{(\textcolor{Maroon}{\downarrow\! 10.5})}$  & \cellcolor[gray]{0.85}$\boldsymbol{16.2}$ $\boldsymbol{(\textcolor{Maroon}{\downarrow\! 7.8})}$  & \cellcolor[gray]{0.85}$\boldsymbol{26.6}$ $\boldsymbol{(\textcolor{Maroon}{\downarrow\! 34.2})}$  & \cellcolor[gray]{0.85}$\boldsymbol{8.3}$ $\boldsymbol{(\textcolor{Maroon}{\downarrow\! 8.9})}$  & \cellcolor[gray]{0.85}$\boldsymbol{14.9}$ $\boldsymbol{(\textcolor{Maroon}{\downarrow\! 13.6})}$  & \cellcolor[gray]{0.85}$\boldsymbol{15.0}$ $\boldsymbol{(\textcolor{Maroon}{\downarrow\! 9.7})}$  & \cellcolor[gray]{0.85}$\boldsymbol{34.3}$ $\boldsymbol{(\textcolor{Maroon}{\downarrow\! 17.0})}$  \\ \hline
                                      & Clean                      & 23.7                         & 9.6                          & 3.0                              & 23.3                         & 15.7                         & 5.6                         & 2.9                              & 23.0                         & 68.6                         \\
                                      & Random                     & 14.1 $(\textcolor{orange}{\downarrow\! 9.6})$  & 6.0 $(\textcolor{orange}{\downarrow\! 3.6})$  & 1.8 $(\textcolor{orange}{\downarrow\! 1.2})$  & 16.9 $(\textcolor{orange}{\downarrow\! 6.4})$  & 10.8 $(\textcolor{orange}{\downarrow\! 4.9})$  & 4.3 $(\textcolor{orange}{\downarrow\! 1.3})$  & 0.9 $(\textcolor{orange}{\downarrow\! 2.0})$  & 20.0 $(\textcolor{orange}{\downarrow\! 3.0})$  & 51.7 $(\textcolor{orange}{\downarrow\! 16.9})$  \\
                                      & $p^*(\boldsymbol{\Theta})$ & 12.5 $(\textcolor{Maroon}{\downarrow\! 11.2})$  & 5.2 $(\textcolor{Maroon}{\downarrow\! 4.4})$  & 1.2 $(\textcolor{Maroon}{\downarrow\! 1.8})$  & 18.0 $(\textcolor{Maroon}{\downarrow\! 5.3})$  & 7.8 $(\textcolor{Maroon}{\downarrow\! 7.9})$  & 3.6 $(\textcolor{Maroon}{\downarrow\! 2.0})$  & 0.6 $(\textcolor{Maroon}{\downarrow\! 2.3})$  & 17.2 $(\textcolor{Maroon}{\downarrow\! 5.8})$  & 43.4 $(\textcolor{Maroon}{\downarrow\! 25.2})$  \\
\multirow{-4}{*}{GPT-4o-mini~\cite{gpt-4o}}         & \cellcolor[gray]{0.85}$\boldsymbol{\Theta}^*$    & \cellcolor[gray]{0.85}$\boldsymbol{9.6}$ $\boldsymbol{(\textcolor{Maroon}{\downarrow\! 14.1})}$  & \cellcolor[gray]{0.85}$\boldsymbol{4.2}$ $\boldsymbol{(\textcolor{Maroon}{\downarrow\! 5.4})}$  & \cellcolor[gray]{0.85}$\boldsymbol{1.6}$ $\boldsymbol{(\textcolor{Maroon}{\downarrow\! 1.4})}$  & \cellcolor[gray]{0.85}$\boldsymbol{16.3}$ $\boldsymbol{(\textcolor{Maroon}{\downarrow\! 7.0})}$  & \cellcolor[gray]{0.85}$\boldsymbol{7.3}$ $\boldsymbol{(\textcolor{Maroon}{\downarrow\! 8.4})}$  & \cellcolor[gray]{0.85}$\boldsymbol{2.9}$ $\boldsymbol{(\textcolor{Maroon}{\downarrow\! 2.7})}$  & \cellcolor[gray]{0.85}$\boldsymbol{0.4}$ $\boldsymbol{(\textcolor{Maroon}{\downarrow\! 2.5})}$  & \cellcolor[gray]{0.85}$\boldsymbol{15.5}$ $\boldsymbol{(\textcolor{Maroon}{\downarrow\! 7.5})}$  & \cellcolor[gray]{0.85}$\boldsymbol{42.7}$ $\boldsymbol{(\textcolor{Maroon}{\downarrow\! 25.9})}$  \\ \hline
                                      & Clean                      & 51.2                         & 14.2                         & 9.7                              & 25.0                         & 40.7                         & 10.9                        & 11.7                             & 24.9                         & 75.4                         \\
                                      & Random                     & 36.5 $(\textcolor{orange}{\downarrow\! 14.7})$  & 10.1 $(\textcolor{orange}{\downarrow\! 4.1})$  & 5.1 $(\textcolor{orange}{\downarrow\! 4.6})$  & 19.1 $(\textcolor{orange}{\downarrow\! 5.9})$  & 24.7 $(\textcolor{orange}{\downarrow\! 16.0})$  & 8.6 $(\textcolor{orange}{\downarrow\! 2.3})$  & 9.5 $(\textcolor{orange}{\downarrow\! 2.2})$  & 20.9 $(\textcolor{orange}{\downarrow\! 4.0})$  & 56.3 $(\textcolor{orange}{\downarrow\! 19.1})$  \\
                                      & $p^*(\boldsymbol{\Theta})$ & 26.4 $(\textcolor{Maroon}{\downarrow\! 24.8})$  & 9.0 $(\textcolor{Maroon}{\downarrow\! 5.2})$  & 4.6 $(\textcolor{Maroon}{\downarrow\! 5.1})$  & 17.7 $(\textcolor{Maroon}{\downarrow\! 7.3})$  & 19.1 $(\textcolor{Maroon}{\downarrow\! 21.6})$  & 6.5 $(\textcolor{Maroon}{\downarrow\! 4.4})$  & 6.5 $(\textcolor{Maroon}{\downarrow\! 5.2})$  & 17.0 $(\textcolor{Maroon}{\downarrow\! 7.9})$  & 50.7 $(\textcolor{Maroon}{\downarrow\! 24.7})$  \\
\multirow{-4}{*}{GPT-4o~\cite{gpt-4o}}              & \cellcolor[gray]{0.85}$\boldsymbol{\Theta}^*$    & \cellcolor[gray]{0.85}$\boldsymbol{20.3}$ $\boldsymbol{(\textcolor{Maroon}{\downarrow\! 30.9})}$  & \cellcolor[gray]{0.85}$\boldsymbol{7.3}$ $\boldsymbol{(\textcolor{Maroon}{\downarrow\! 6.9})}$  & \cellcolor[gray]{0.85}$\boldsymbol{4.7}$ $\boldsymbol{(\textcolor{Maroon}{\downarrow\! 5.0})}$  & \cellcolor[gray]{0.85}$\boldsymbol{16.1}$ $\boldsymbol{(\textcolor{Maroon}{\downarrow\! 8.9})}$  & \cellcolor[gray]{0.85}$\boldsymbol{15.8}$ $\boldsymbol{(\textcolor{Maroon}{\downarrow\! 24.9})}$  & \cellcolor[gray]{0.85}$\boldsymbol{5.5}$ $\boldsymbol{(\textcolor{Maroon}{\downarrow\! 5.4})}$  & \cellcolor[gray]{0.85}$\boldsymbol{5.4}$ $\boldsymbol{(\textcolor{Maroon}{\downarrow\! 6.3})}$  & \cellcolor[gray]{0.85}$\boldsymbol{14.6}$ $\boldsymbol{(\textcolor{Maroon}{\downarrow\! 10.3})}$  & \cellcolor[gray]{0.85}$\boldsymbol{50.3}$ $\boldsymbol{(\textcolor{Maroon}{\downarrow\! 25.1})}$  \\ \hline
\end{tabular}}
\vspace{-0.3cm}
\label{exp:3}
\end{table*}

\subsection{Experimental Setup}
\noindent\textbf{Tasks \& Datasets.} We evaluate VLMs' robustness to 3D variations using AdvDreamer on three representative vision-language tasks. (1) For \emph{zero-shot image classification}, we utilize the ImageNet test set \cite{deng2009imagenet}. (2) For \emph{image captioning}, we employ the test splits from COCO Caption \cite{chen2015microsoft} and NoCaps \cite{agrawal2019nocaps} datasets. (3) For \emph{Visual Question Answering (VQA)}, we use the VQAv2 \cite{goyal2017making} test set. For each dataset, we select 300 clean samples spanning 30 categories with clear and well-defined object instances. During the optimization process, we employ ImageNet-1K categories as the semantic label set $\mathcal{Y}$ for the classification task and use COCO-80 categories for the captioning and VQA tasks.

\noindent\textbf{Victim Models.} For zero-shot classification, we target vision-language foundation models: OpenCLIP~\cite{cherti2023reproducible,openclip}, BLIPs~\cite{li2022blip, li2023blip}, and additional models for transfer attacks. For other tasks, we select mainstream open-source VLMs (LLaVa~\cite{liu2024visual,liu2024improved}, MiniGPT-4~\cite{zhu2023minigpt}, \etc) as well as representative commercial VLMs (GPT-4o and GPT-4o-mini).

\noindent\textbf{Methods \& Baselines.} We use two variants with AdvDreamer: (1) $p^*(\boldsymbol{\Theta})$: represents the average results over 10 Adv-3DT samples randomly sampled from the optimal adversarial distribution, while (2) $\boldsymbol{\Theta}^*$: indicates the evaluation results using the Adv-3DT sample corresponding to the distribution center. Additionally, we establish two potential baselines following \cite{dong2022viewfool, ruan2023towards}: the original natural images without transformation (\textbf{\emph{Clean}}) and the average results over 10 images with random 3D transformations (\textbf{\emph{Random}}).

\noindent\textbf{Metrics.} For zero-shot classification, we follow the CLIP-Benchmark~\cite{CLIP-Benchmark} protocol, reporting Top-1 accuracy and its degradation relative to clean samples. For captioning, in addition to conventional metrics (CIDEr, SPICE, BLEU), we adopt the LLM-as-a-judge approach~\cite{liu2024visual,zheng2023judging,zhang2024benchmarking} to address the open-ended nature of VLM responses, aligning evaluations more closely with human preferences. We use GPT-4 to assess three aspects of the response on a 10-point scale: semantic accuracy, tone confidence, and overall coherence, with the scores summed to yield the \textbf{\emph{GPT-Score}} (see Fig.\ref{fig:vis} (D)). For the VQA task, we introduce \textbf{\emph{GPT-Acc}}, which leverages LLM-based judgment to evaluate response correctness. The judgment prompts are detailed in Appendix\textcolor{iccvblue}{C}.



\subsection{Research Questions (RQs) and Findings}


\noindent\textbf{\colorbox{elegantblue}{RQ1:} Do current VLMs demonstrate sufficient robustness to 3D variations?} \emph{\textbf{Take-away:} "Emphatically not. Our evaluation reveals the severe vulnerability of current VLMs to worst-case 3D variations across multiple vision-language tasks."} As shown in Tab.~\ref{exp:1}, in zero-shot classification, Adv-3DT samples generated by AdvDreamer significantly compromise VLMs' performance, causing accuracy drops of up to 80\%. This substantially outperforms the Random baseline, validating AdvDreamer's effectiveness in capturing worst-case transformations. Notably, samples under $\boldsymbol{\Theta}^*$ induce more severe performance degradation than those under the adversarial distribution $p^*(\boldsymbol{\Theta})$, indicating the existence of extremely vulnerable points.

This fundamental vulnerability extends beyond classification to more complex open-ended tasks. As illustrated in Tab.\ref{exp:3}, for image captioning, Adv-3DT samples under both $p^*(\boldsymbol{\Theta})$ and $\boldsymbol{\Theta}^*$ effectively degrade model performance across all evaluation metrics, with GPT-Score decreasing by nearly 50\% across various VLMs. This significant degradation reveals a critical limitation in current VLMs' 3D understanding and reasoning capabilities. For VQA tasks, GPT-4o, despite achieving the highest accuracy (75.4\%) on clean samples—exhibits a substantial 25\% accuracy drop when under Adv-3DT attacks. Fig.~\ref{fig:vis} (A) provides visualizations of representative Adv-3DT samples generated by AdvDreamer, with more examples presented in Appendix.\textcolor{iccvblue}{D}.


\noindent\textbf{\colorbox{elegantblue}{RQ2:} To what extent do different VLMs share common 3D variation vulnerability regions?} \emph{\textbf{Take-away:} "Substantially. Our analysis reveals that Adv-3DT samples demonstrate strong cross-model transferability across various VLMs, indicating shared vulnerability regions in the 3D variation space."} As demonstrated in Tab.~\ref{exp:2}, we investigate transferability by utilizing OpenCLIP and BLIP as surrogate models to optimize $p^*(\boldsymbol{\Theta})$, then evaluate six target VLMs with diverse architectures and training objectives. The transfer attacks maintain strong effectiveness, with performance degradation comparable to direct attacks. Notably, $p^*(\boldsymbol{\Theta})$ exhibits smaller transfer gaps than $\boldsymbol{\Theta}^*$, suggesting that learning the underlying distribution can better capture generalizable vulnerability regions. This transferability extends consistently to captioning and VQA tasks (Tab.~\ref{exp:3}). When employing OpenAI-CLIP as the surrogate encoder, Adv-3DT samples generated by AdvDreamer effectively compromise commercial systems like GPT-4o. We attribute this universal transferability to a fundamental limitation: current image-text pretraining datasets exhibit significant 3D biases and fail to encompass diverse certain special 3D transformations, leading to universal performance deficiencies under these transformation regions.
\begin{table}[t]
\caption{Top-1 Accuracy and Naturalness Score of Adv-3DT samples generated by AdvDreamer w/ and w/o NRM. (under $p^*(\boldsymbol{\Theta})$)}
\vspace{-0.2cm}
\setlength\tabcolsep{10.0pt}
\renewcommand\arraystretch{1.0}
\centering
\scalebox{0.7}{
\begin{tabular}{c|c|c|c}
\hline
  Methods                  & Top-1 Acc.  & $Score_R$ & $Score_P$ \\ \hline
AdvDreamer w/o. NRM & \textbf{48.6}\% $\boldsymbol{(\textcolor{Maroon}{\downarrow\!49.4})}$  & 1.60     & 1.39    \\
AdvDreamer w/. NRM  & 54.0\% $(\textcolor{Maroon}{\downarrow\! 44.0})$& \textbf{2.52}    & \textbf{2.57}    \\ \hline
\end{tabular}}
\vspace{-0.3cm}
\label{NRM2}
\end{table}
\begin{figure}[t]
  \centering
  \includegraphics[height=3.0cm]{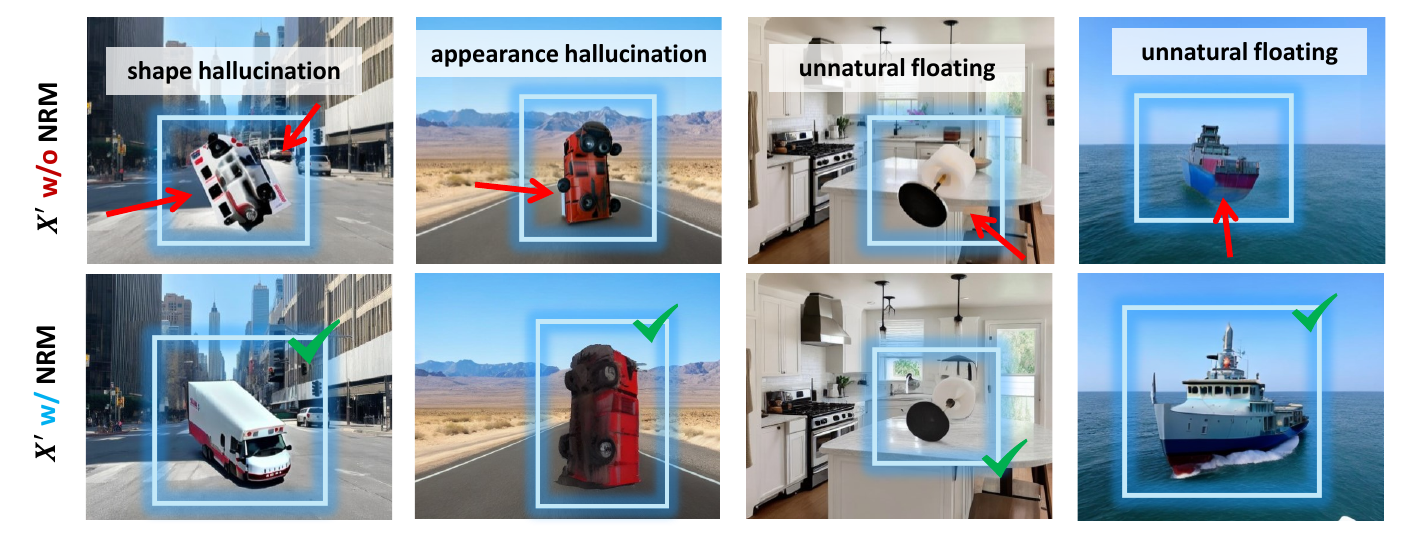}
  \vspace{-0.2cm}
  \caption{Visualization of Adv-3DT samples w/ and w/o NRM.}
  \vspace{-0.6cm}
  \label{fig:failure-sample}
\end{figure}
\noindent\textbf{\colorbox{elegantblue}{RQ3:} Does the performance degradation under Adv-3DT samples generated by AdvDreamer fundamentally arise from 3D variations?} \emph{\textbf{Take-away:} "Yes, we successfully reproduce these Adv-3DT samples in physical environments and demonstrate that degradation is inherently caused by the 3D variation rather than undesirable image quality problems."} With the naturalness feedback provided by NRM, AdvDreamer effectively prevents unwanted degradation of samples during the optimization process. To verify this, we evaluate Adv-3DT samples generated by AdvDreamer with and without NRM under ImageNet. Tab.~\ref{NRM2} presents the accuracy under OpenCLIP (ViT-B/16), alongside GPT-4o-assessed visual realism ($Score_R$) and physical plausibility ($Score_P$). The results indicate that NRM 
significantly improves samples' realism and physical feasibility, with a marginal trade-off in aggressiveness. Furthermore, Fig.~\ref{fig:failure-sample} presents a qualitative comparison, illustrating the effectiveness of NRM's naturality feedback in improving the image quality and mitigating the distribution discrepancy compared to natural image distributions. In addition, we validate the alignment between NRM predictions and human evaluations, as detailed in Appendix~\textcolor{iccvblue}{C.2}.

\begin{table}[t]
\caption{Quantitative results of physical experiments.}
\vspace{-0.1cm}
\setlength\tabcolsep{11.8pt}
\renewcommand\arraystretch{1.0}
\centering
\scalebox{0.65}{
\begin{tabular}{l|cc|cc}
\hline
\multirow{3}{*}{Sample} & \multicolumn{2}{c|}{\multirow{2}{*}{\begin{tabular}[c]{@{}c@{}}Zero-shot Cls. \\ (OpenCLIP-B)\end{tabular}}} & \multicolumn{2}{c}{\multirow{2}{*}{\begin{tabular}[c]{@{}c@{}}VQA \\ (LLaVa-1.5-7B)\end{tabular}}} \\
                        & \multicolumn{2}{c|}{}                                                                                               & \multicolumn{2}{c}{}                                                                               \\ \cline{2-5} 
                        & Acc.                                                      & Conf.                                                    & Acc.1                                            & Acc.2                                             \\ \hline
Natural                 & 100.0                                                    & 0.892                                                    & 83.3                                            & 100.0                                            \\ \hline
Adv-3DT (AdvDreamer)     & 0.0~\scriptsize{$(\textcolor{Maroon}{\downarrow\! 100})$}                                                     & 0.003                                                    & 25.0~\scriptsize{$(\textcolor{Maroon}{\downarrow\! 58})$}                                            & 25.0~\scriptsize{$(\textcolor{Maroon}{\downarrow\! 75})$}                                             \\ \rowcolor{gray!25}
Adv-3DT (Physicial)      & 51.3~\scriptsize{$(\textcolor{Maroon}{\downarrow\! 49})$}                                                     & 0.413                                                    & 33.6~\scriptsize{$(\textcolor{Maroon}{\downarrow\! 50})$}                                            & 45.1~\scriptsize{$(\textcolor{Maroon}{\downarrow\! 55})$}                                             \\ \hline
\end{tabular}
}
\vspace{-0.65cm}
\label{exp:4}
\end{table}
Furthermore, we conduct physical-world experiments on 12 common objects from traffic and household environments to validate this. Our experimental procedure consisted of three main steps: \emph{\textbf{1)}} capturing images of objects in their natural state, \emph{\textbf{2)}} using AdvDreamer with OpenCLIP as the surrogate model to generate Adv-3DT samples, and \emph{\textbf{3)}} physically reproducing the adversarial 3D transformations captured by AdvDreamer through video recording. This process yielded 3,807 frames of physical Adv-3DT samples. We evaluate effectiveness through zero-shot classification and VQA tasks. For VQA, we designed two evaluation protocols: multiple-choice classification (Acc.1) and binary verification (Acc.2). As shown in Table~\ref{exp:4}, physically reproduced Adv-3DT samples significantly degraded VLM performance: the accuracy of zero-shot classification dropped to 51.3\%, while VQA performance decreased to 33.6\% (Acc.1) and 45.1\% (Acc.2). These results confirm that AdvDreamer successfully captures physically realizable 3D variations that represent real world corner cases. However, an aggressiveness gap between digital and physical attacks is observed, primarily attributed to data distribution differences and camera parameter fluctuations. Fig.~\ref{fig:vis} (B) presents representative examples from physical experiments, with more detailed results provided in Appendix~\textcolor{iccvblue}{D}.

\begin{figure}[t]
  \centering
  \includegraphics[height=2.9cm]{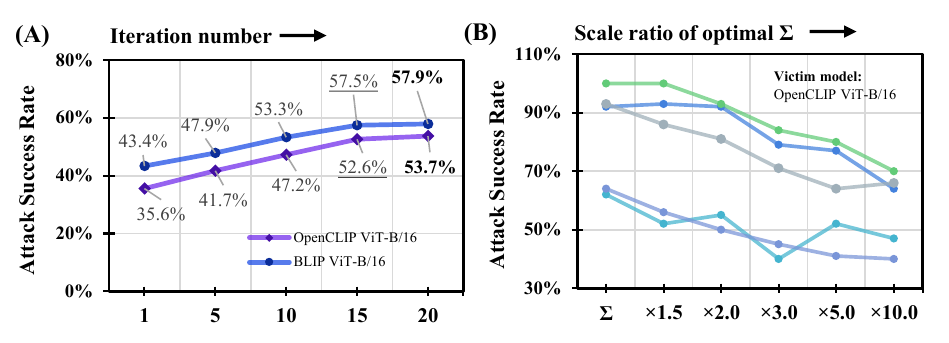}
    \vspace{-0.3cm}
  \caption{Attack success rate of Adv-3DT samples \wrt (A) optimization steps and (B) scaling ratios of the optimal distribution.}
    \vspace{-0.5cm}
  \label{fig:abla}
\end{figure}

\subsection{Ablation Studies and Additional Analysis} \label{sec:ablation}


\noindent\textbf{Comparison of $\mathcal{L}_\text{ISP}$ with Alternative Loss Forms}: We further compared three alternative loss forms in~\cite{cui2024robustness, zhao2023evaluate}: matching text-text features ($MF_{tt}$), matching image-image features ($MF_{ii}$) and matching image-text features ($MF_{it}$), all of which are implemented using image-text alignment embeddings. Indeed, $\mathcal{L}_{\text{ISP}}$ can be viewed as an probabilistic variant of $MF_{it}$. As shown in Tab.~\ref{loss}, on OpenCLIP (ViT-B/16), $\mathcal{L}_\text{ISP}$  achieves the highest ASR on $p^*(\boldsymbol{\Theta})$ while maintaining relatively high ASR on $\boldsymbol{\Theta}$. Though combining $MF_{ii}$ and $MF_{tt}$ provides a slight improvement, their text-image modality switching process introduces extra computational overhead. Thus, we consider $\mathcal{L}_\text{ISP}$ the optimal choice for balancing performance and efficiency.

\noindent\textbf{Convergence Analysis \& Computational Efficiency.} We analyze the attack success rates (ASR) of $p^*(\boldsymbol{\Theta})$ against OpenCLIP (ViT-B/16) and BLIP (ViT-B/16) across different optimization iterations. As shown in Fig.~\ref{fig:abla}, AdvDreamer's effectiveness consistently improves with more iterations and converges after 15 steps. Thus, we set the iteration step default to 15 throughout our experiments to balance effectiveness and efficiency. The computational cost is discussed in Appendix~\textcolor{iccvblue}{C.4} by detailing the overhead of each component. For each clean sample, AdvDreamer requires an average of 0.28 GPU hours with a single RTX 3090.

\noindent\textbf{Does Optimal Distribution $p^*(\boldsymbol{\Theta})$ Effectively Characterize the Worst-case 3D Variations?} We conduct analysis on ImageNet using five random natural images. For each image, we generate 100 Adv-3DT samples by scaling the variance $\boldsymbol{\Sigma}$ of their $p^*(\boldsymbol{\Theta})$ with different ratios. Fig.~\ref{fig:abla} (B) shows that ASR consistently decreases as the sampling range expands beyond the optimal distribution. This inverse relationship confirms that AdvDreamer successfully identifies the most vulnerable regions in the 3D variation space.


\noindent\textbf{Background Variations.} While generalizable scene representation remains a significant challenge, we explore an alternative approach by examining the effects of combining transformed foregrounds with various backgrounds. As shown in Fig.~\ref{fig:vis} (C), our qualitative results demonstrate that Adv-3DT samples generated by AdvDreamer effectively generalize across diverse potential scenes.

\begin{table}[t]
\caption{Top-1 Accuracy and Degradation (\textcolor{Maroon}{$\downarrow$\%}) of Adv-3DT samples generated by AdvDreamer with alternative loss forms}
\vspace{-0.2cm}
\setlength\tabcolsep{11.0pt}
\renewcommand\arraystretch{1.0}
\centering
\scalebox{0.67}{
\begin{tabular}{c|c|c}
    \hline
       Loss form         & $p^*(\boldsymbol{\Theta})$ & $\boldsymbol{\Theta}$    \\ \hline
    $MF_{it}$ (Equivalently in~\cite{cui2024robustness,zhao2023evaluate})    & 58.1\% ($\textcolor{Maroon}{\downarrow\! 39.9}$)  & 22.9\% ($\textcolor{Maroon}{\downarrow\! 75.1}$) \\
    $MF_{ii}$~\cite{zhao2023evaluate}      & 56.2\% ($\textcolor{Maroon}{\downarrow\! 41.8}$)  & 22.2\% ($\textcolor{Maroon}{\downarrow\! 75.8}$) \\
    $MF_{ii}$+$MF_{tt}$~\cite{zhao2023evaluate} & \underline{54.9\%} ($\textcolor{Maroon}{\downarrow\! 43.1}$)  & \textbf{17.1\%} ($\boldsymbol{\textcolor{Maroon}{\downarrow\! 80.9}}$) \\
    \rowcolor{gray!15} $\mathcal{L}_{\text{ISP}}$ (Ours)     & \textbf{54.0\%} ($\boldsymbol{\textcolor{Maroon}{\downarrow\! 44.0}}$)  & \underline{17.4\%} ($\textcolor{Maroon}{\downarrow\! 80.6}$) \\ \hline
    \end{tabular}}
\vspace{-0.3cm}
\label{loss}
\end{table}
\subsection{MM3DTBench}
\begin{figure}[t]
  \centering
  \includegraphics[height=2.4cm]{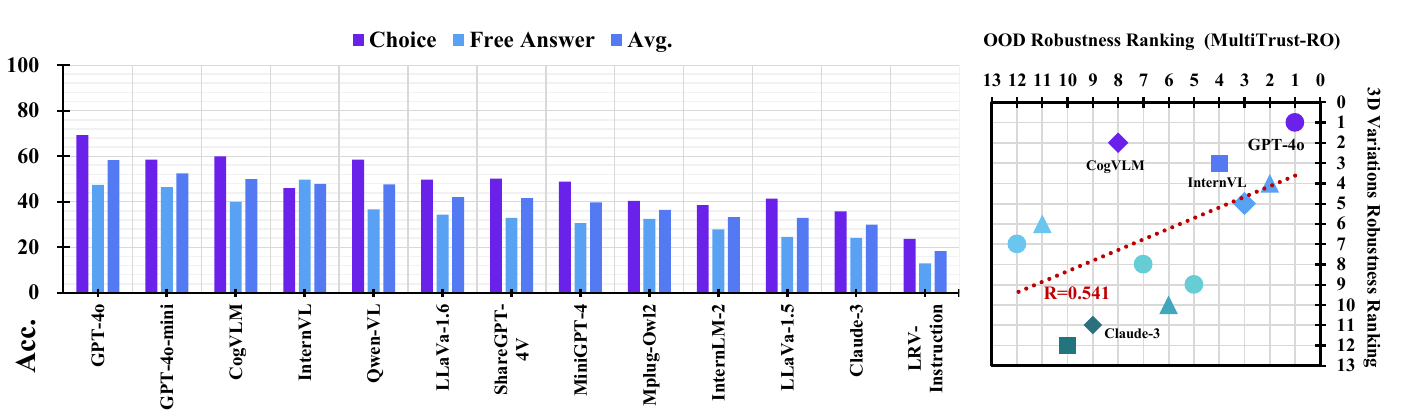}
  \vspace{-0.2cm}
  \caption{Benchmarking result of VLMs' 3D variation robustness.}
  \vspace{-0.6cm}
  \label{fig:bench}
\end{figure}
Finally, We introduce MM3DTBench, it comprises 215 challenging Adv-3DT samples, consisting of samples generated by AdvDreamer and their physical replications. For each sample, we provide question templates with candidate answers and precise orientation annotations for the transformed objects. We employ two evaluation tasks: multiple-choice (\textbf{\emph{Choice}}) and free-form description (\textbf{\emph{Free answer}}), with accuracy computed through text-only GPT-4 judgment. More Details are provided in the Appendix.\textcolor{iccvblue}{E}. Our evaluation of \textbf{\emph{13}} representative VLMs reveals robustness gaps. As shown in Fig.~\ref{fig:bench}, even top-performing models (GPT-4o~\cite{gpt-4o}, CogVLM~\cite{wang2023cogvlm}, and Intern-VL~\cite{chen2024internvl}) achieve limited success, while most models struggle with accuracy below 50\%. Notably, Claude-3~\cite{claude-3}, despite its recognized excellence, only achieves an accuracy of 30.0\%. This performance correlates with models' general out-of-distribution robustness measured by MultiTrust~\cite{zhang2024benchmarking}, suggesting a fundamental limitation of current VLMs. We encourage the adoption of MM3DTBench in VLM development to advance security evaluation in dynamic scenarios.

\section{Conclusion}
We proposed AdvDreamer, a novel framework for characterizing real-world adversarial 3D transformation samples from single-view observations. Our comprehensive evaluation revealed critical robustness gaps in existing VLMs, highlighting the urgent need to enhance VLMs' 3D variation perception and understanding capabilities in safety-critical applications. Through MM3DTBench, we established the first benchmark for worst-case 3D variations, aiming to facilitate the development of more robust vision-language systems for real-world deployment.

\clearpage
{
    \small
    \bibliographystyle{ieeenat_fullname}
    \bibliography{main}
}

\end{document}